\def\year{2017}\relax
\documentclass[letterpaper]{article}
\usepackage{aaai17}
\usepackage{times}
\usepackage{helvet}
\usepackage{courier}

\usepackage{graphicx}
\usepackage{balance}
\usepackage{subcaption}
\usepackage{graphics} 

\usepackage{amsmath}

\usepackage{subcaption}
\usepackage{multirow}

\begin{document}

\title{A World of Difference: Divergent Word Interpretations among People}

\author{%
	  Tianran Hu$^{*1}$, Ruihua Song$^{\dag2}$, Maya Abtahian$^{*3}$, Philip Ding$^{\dag4}$, Xing Xie$^{\dag5}$, Jiebo Luo$^{*6}$\\
	  	$^*$University of Rochester, $^\dag$Microsoft Research\\
	  \{$^1$thu, $^6$jluo\}@cs.rochester.edu\\
      $^3$maya.r.abtahian@rochester.edu\\
	  \{$^2$song.ruihua,$^4$phding,$^5$xing.xie\}@microsoft.com
}

\maketitle

\begin{abstract}
Divergent word usages reflect differences among people. In this paper, we present a novel angle for studying word usage divergence -- word interpretations. We propose an approach that quantifies semantic differences in interpretations among different groups of people. The effectiveness of our approach is validated by quantitative evaluations. Experiment results indicate that divergences in word interpretations exist. We further apply the approach to two well studied types of differences between people -- gender and region. The detected words with divergent interpretations reveal the unique features of specific groups of people. For gender, we discover that certain different interests, social attitudes, and characters between males and females are reflected in their divergent interpretations of many words. For region, we find that specific interpretations of certain words reveal the geographical and cultural features of different regions. 
\end{abstract}

\section{Introduction}
As Pinker stated in his book~\cite{pinker2007stuff}, \textit{``language is the window into human nature''}. The differences among people, for example, gender~\cite{coates2015women}, age~\cite{nguyen2013old}, and occupation~\cite{hu2016language} trigger their different word usages. These divergences in word usages, in turn, reveal the unique features of groups of people, such as their specific interests~\cite{rakesh2014personalized}, personalities~\cite{vinciarelli2014survey}, 
and so on. The previous work in differences among people mostly focuses on word frequency. In this work, we take a step further to study word interpretation. Our findings suggest that the interpretations of words may vary from people to people. More importantly, we discover that divergent word interpretations are also related to the unique features of people, which provides a novel angle to comprehend the differences between social groups.


Intuitively, a word can be interpreted using its semantically close words in a corpus. The idea of our approach is to extract words' semantically closest words as their interpretations from corpora of different groups of people, separately. The widely adopted word embedding techniques are employed to learn the closest words. We then quantify the semantic distance between the multiple interpretations of each word. The larger the distances is, the more differently a word is interpreted among people. We refer to the semantic distance between the interpretations of a word as its \textit{\textbf{divergence score}}, and the words of high divergence scores as \textit{\textbf{divergent words}}.

We apply the approach to studying two widely studied differences among people -- gender and regional differences. The discovered divergent words clearly portray the unique features of specific populations. For gender, we observe the different interests between men and women in various aspects. As for the regionally divergent words, we observe that they are usually related to the geographical or cultural features of regions. In this paper, we report the detected gender and regionally divergent words, and summarize the unique features of populations as read from these words.

\section{Related Work}

Much work has focused on divergent word usages among people in numerous aspects, such as gender, age, occupation, and region. It is reported that typical male language uses more judgmental adjectives, elliptical sentences, directives, and ``I'' reference,  while typical female language contains more intensive adverbs, references to emotions, uncertainty verbs~\cite{poynton1989language}. People in different age groups appear to choose words differently. Nguyen et al. reported that, comparing with younger people, older people talk more about family and work, and use less swear words in their language~\cite{nguyen2013old}. 
Jobs affect the language patterns of people as well~\cite{hu2016language}. In the paper, the authors listed the most used words of several occupations, and their results indicate clear divergences in word usages between people of different jobs. Similarly, different cities may have different preferences in words~\cite{cheng2010you}, which reflect regional differences between people.

\section{Data Collection and Preprocessing}
We collect our dataset via the Twitter open APIs\footnote{https://dev.twitter.com/overview/api}. First, we set up two geo-bounding boxes. Both boxes are two degrees of latitude long, and two degrees of longitude wide. One box is centered at NYC, and the other is centered at the Bay Area. We use the Twitter streaming API to collect users who have posted tweets within these two areas from June to October 2016, resulting in 0.4 million unique users. For each unique user, we download their at most 3,000 recent tweets with the Twitter timeline API, resulting in around 0.2 billion tweets. 

To assign gender tags to each user, we employ the API of genderize.io\footnote{https://genderize.io} suggested in~\cite{abbar2015you}. Genderize.io takes a first name as input, and outputs the probabilities of this name being used by males and females. We feed the API with the first names appearing in user profiles to obtain gender tags, and filter out the names with low confidence (probability $<$ 0.7). We obtain around 32 thousand males, and 30 thousand females. To figure out user locations, we follow the approach suggested in~\cite{sadilek2013modeling}. We collect all the geo-tagged tweets from a user's recent tweets, and filter out the user if the amount of geo-tagged tweets was less than 10, or if less than half of the tweets were sent inside our bounding boxes. This step leaves us with roughly 33 thousand NYC users, and 33 thousand Bay Area users.

We further clean these raw tweets by removing hashtags, mentions, URLs, retweets, and short tweets (less than 10 words). We also exclude the words that are used by few people (less than 100 people) from these tweets, and set all words to lowercase. After cleaning, we separate these tweets according to user gender and location. Therefore, we obtain four corpora: male and female corpora denoted by $C_m$ and $C_f$, contains 7.4 million, and 5.4 million tweets, respectively; and NYC and Bay Area corpora denoted by $C_N$ and $C_B$, contains 8.1 million and 7.9 million tweets, respectively.

\section{Methodology}
Our approach consists of two steps. First, it learns group specific interpretations for each word. Second, it quantifies the semantic distance between the interpretations. Without loss of generality, we denote the corpora of two groups of people as $C_1$ and $C_2$. These groups of people can be people of different genders, from different regions, or of any other type of difference, and our approach can be easily extended to more than two groups.

\subsection{Group Specific Word Interpretations}

To learn the word interpretations for two groups of people, we feed their corpora to an embedding model, and obtain a word vector space for each group. Hence, using $C_1$ and $C_2$, we train two word vector spaces $S_1$ and $S_2$, respectively. Each space contains the semantic patterns of a group. Since words' closest words in an embedding space provide effective clues to understanding the words, we take these closest words as their interpretations. Therefore, for each word $w$, we collect two sets of its top $n$ closest words from $S_1$ and $S_2$, respectively, along with the semantic similarities between these words to $w$. We refer to such sets as $w$'s \textit{\textbf{Interpreting Sets}}, denoted by $I_{w}^{1}$ and $I_{w}^{2}$. 

\begin{figure}
\centering
\includegraphics[width=0.9\columnwidth]{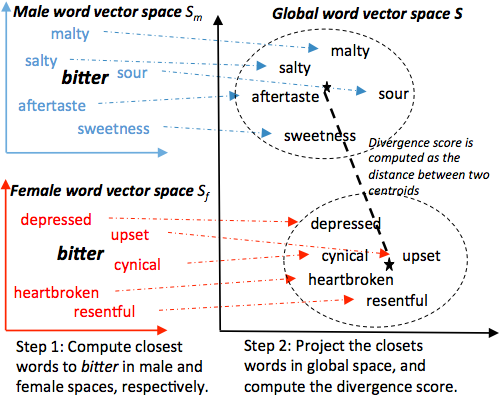}
\caption{Illustration of the calculation of the divergence score of \textit{bitter} from male and female corpora. For the brevity of the illustration, we do not note the similarities between closest words and word \textit{bitter} on the plot.}~\label{fig:space}
\end{figure}

The left part of Figure~\ref{fig:space} shows an example of extracting the interpretations of \textit{bitter} for males and females. In two word vector spaces of the two genders, we find the neighbors of the word, and obtain its two interpreting sets $I_{bitter}^{m}$ and $I_{bitter}^{f}$. For example, set $n = 3$, then the two sets would be as follows:
\begin{align}
I_{bitter}^{m} = \{(salty, 0.7), (sour, 0.6), (aftertaste, 0.6) \}\notag
\\
I_{bitter}^{f} = \{(upset, 0.6), (depressed, 0.6), (cynical, 0.6) \}\notag
\end{align} where the value after each word denotes its semantic similarities to \textit{bitter}. The interpreting sets of a word represent its group specific understandings. If $w$ is interpreted similarly by two groups, $I_{w}^{1}$ and $I_{w}^{2}$ should be similar as well. Otherwise, the difference between two sets reflects the divergence between $w$'s interpretations.  

\subsection{Quantified Interpretation Divergences}


To quantify the interpretation divergence, we first project the words in both interpreting sets to a global word vector space $S$. In this space, we then compute the semantic distance between these two groups of words. The global space $S$ is trained using both corpora ($C_1 + C_2$), and contains the overall language patterns of both groups. To be specific, we find the corresponding word vectors in $S$ for words in $I_{w}^{1}$ and $I_{w}^{2}$, respectively. We then compute the centroids of two groups of words in the global space as the weighted average of the word vectors, denoted by $cent(I_{w}^{1})$ and $cent(I_{w}^{2})$. The similarities between words and $w$ are the averaging weights. These two centroids are used as the representatives of the interpreting sets in $S$, as suggested in~\cite{reisinger2010multi}. The cosine distance between two centroids is then calculated to measure the semantic difference between the interpreting sets. We refer to this distance as $w$'s divergence score between two groups of people. Note that, the calculation in a global space takes the semantic meanings of closest words into account. Meanwhile, the weighted average assigns different importance to the words in interpreting sets.

The right part of Figure~\ref{fig:space} shows an example of quantifying the differences between male and female interpreting sets of \textit{bitter}. Centroids (denoted by asterisks in the figure) of interpreting sets are first computed. For example, $cent(I_{bitter}^{m})$ is an aggregated word vector in $S$, and calculated as:
\begin{align*}
\frac{0.7 \times v(salty) + 0.6 \times v(sour) + 0.6 \times v(aftertaste)}{0.7 + 0.6 + 0.6}
\notag
\end{align*} where $v(salty)$ denotes $salty$'s word vector in $S$. The cosine distance between $cent(I_{bitter}^{m})$ and $cent(I_{bitter}^{f})$ is then computed as \textit{bitter}'s divergence score.

\section{Gender Divergent Words}

We manually check the male and female interpretations (words in two interpreting sets) of the top 100 gender divergent words, in terms of divergent score. We observe that people of one gender are more likely to relate words to the specific interests of this gender.



\subsubsection{Sports}
Males are generally more interested in sports than females~\cite{deaner2015sex}. This conclusion can be derived from the observation that many words associated with sports by men are interpreted quite differently by women. Such words include \textit{title}, \textit{finals}, \textit{draft}, and many others. The closest words to \textit{title} for males are \textit{championship}, \textit{contender}, and \textit{undefeated}. This interpretation indicates that males are more likely to connect the word with championship titles. Differently, female closest words to the word are \textit{draft}, \textit{rewrite}, and \textit{novel}, implying article and book titles. \textit{Finals} is related to the final tournaments in sport for males (closest words: \textit{playoffs}, \textit{tourney}, and \textit{semifinals}). For females, it refers to final exams (closest words: \textit{midterms}, \textit{midterm}, and \textit{assignments}). Males are more like to connect the word \textit{draft} with the player selection procedures of professional sport teams, with closest words such as \textit{drafted}, \textit{rounders}, and \textit{76ers}. For females, the word is closest to \textit{manuscript}, \textit{revision}, and \textit{outline}, implying literary drafts.


\subsubsection{Fashion}
Females are more into fashion than males, as reported in~\cite{goldsmith1999fashion}. The word \textit{styles}, for example, is more related to dressing styles for females. They connect this word with \textit{nail}, \textit{hair}, \textit{lipsticks}, and \textit{color}. Different for males, this word is more related to design styles, and closest to \textit{themes}, \textit{designs}, and \textit{typefaces}. The difference also reflects in their interpretations of \textit{shoots}. Females are more likely to use the word to mean shooting pictures, with closest words such as \textit{posing}, \textit{photoshoot}, and \textit{boudoir}. Males refer to this word as firing guns (closest words: \textit{kills}, \textit{shooting}, \textit{fatally}). Another example is \textit{navy}. Females interpret this word as a style and color, and tend to use dressing related terms to interpret it, such as \textit{collar}, \textit{vest}, \textit{berets}, and \textit{striped}. Not surprisingly, males relate this word to military (closest words: \textit{seal}, \textit{marine}, and \textit{airforce}).

\subsubsection{IT \& Video Games}
According to previous work~\cite{gefen1997gender}, men show more interest in IT than women, and many gender divergent words provide evidence of it. An obvious example is \textit{windows}. Males tend to refer to the word as the operating system introduced by Microsoft, and its closest words include names of other operating systems such as \textit{win8}, \textit{win7}, \textit{ubuntu}, and \textit{linux}. For females, \textit{windows} is more related to the architectural structure, and closest to \textit{sunroof}, \textit{doors}, and \textit{curtains}. \textit{Bugs} is also assigned divergent interpretations by two genders. It stands for software bugs for male (closest words: \textit{fixes}, \textit{kernel}, \textit{xcode}, and \textit{glitches}), while it is associated with insects by females (closest words: \textit{mosquitoes}, \textit{ants}, \textit{rodents}, and \textit{spiders}). 

Men's enthusiasm for video games is clearly reflected in their interpretations of some words that relate to this topic. \textit{Destiny}, for example, is related to the famous online game for male, since its closest words are \textit{halo} (video game series), \textit{bungie} (vedio game company), and \textit{xbox} (video game console). Similarly, word \textit{steam} is associated with the video game digital distribution software by males, indicated by closest words such as \textit{valve} (video game company that developed Steam), \textit{xbox}, and \textit{hearthstone} (online game). The explanations of these two words for females are not related to video games at all. Female closest words to \textit{destiny} are \textit{fate}, \textit{greatness}, and \textit{greatness}, and the closest words to \textit{steam} are \textit{compressor}, \textit{dust}, and \textit{heating}.

\section{Regionally Divergent Words}
In this section, we discuss words with divergent interpretations between NYC and the Bay Area. The differences between regions such as lifestyles~\cite{hu2016tales}, eating habits, and cultures~\cite{silva2014you} are reported in many studies. In this work, we also manually check the top 100 regionally divergent words. We observe that there are mainly two types of regionally divergent words: one type reflects geographical differences, and the other reflects the cultural differences between regions. 

\subsection{Geographical Differences between Regions}
Place names account for a large part of this type of divergent words. Such words include \textit{queens} (Queens Borough, NYC), \textit{Jose} (San Jose City, Bay Area), \textit{castro} (Castro District, Bay Area), \textit{buffalo} (Buffalo City, NY state), and so on. An interesting example of this type is \textit{berkeley}. 
In NYC, the word is referred to as the famous university, and closest to other famous universities such as \textit{princeton}, \textit{cornell}, \textit{dartmouth}, and \textit{columbia}. This word is more likely to be interpreted as the Berkeley City in the Bay Area with closest words such as \textit{pasadena}, \textit{rockridge}, \textit{cerrito} (all three are place names in the Bay Area). 

There are also words, which although are not place names, are assigned geographical interpretations in certain regions. \textit{Bart}, meaning Bay Area Rapid Transit in the Bay Area, is a typical example. Its closest words in the Bay Area are \textit{train}, \textit{muni} (Muni Transit), \textit{subway}. In NYC, it is more referred to as a first name. Interestingly, its interpreting set contains \textit{marge}, \textit{homer}, \textit{lisa}, and so on (Bart, Homer, Marge, Lisa are members of the Simpson family in the popular cartoon TV show). 

\subsection{Cultural Differences between Regions}
More importantly, some regionally divergent words reveal the unique cultural features of regions. We summarize the features in several aspects, such as IT, finance, and so on.

\subsubsection{IT} 
The engineering culture in the Bay Area is clearly reflected in the observations that people there are more likely to relate many words to technologies. Such words include \textit{react}, \textit{code}, \textit{port}, and many others. \textit{React} in the Bay Area is interpreted as the reacting of software, indicated by closest words such as \textit{webpack}, \textit{filesystem}, and \textit{objc} (short for Objective-C). In NYC, the word is interpreted literally with closest words such as \textit{respond}, \textit{confuse}, \textit{reacting}, and so on. \textit{Code}, not surprisingly, in the Bay Area is associated with programming, as interpreted by technical terms such as \textit{compiler}, \textit{parser}, \textit{regex}, and \textit{html}. In NYC, this word is more related to coupon codes or discount codes (closest words: \textit{coupon}, \textit{discount}, and \textit{promocode}). As to \textit{port}, this word means harbor in NYC, with closest words such as \textit{terminal}, \textit{ferry}, and \textit{dock}. In the Bay Area, although one of the closest word in its interpreting set is \textit{tacoma} (Tacoma Port, the Bay Area), its other closest words are all related to hardware ports such as \textit{ethernet}, \textit{connector}, and \textit{eero} (Eero Inc., a Wi-Fi company).

\subsubsection{Fashion \& Arts}
Word interpretations in NYC, one of the world's great fashion and art capitals, are biased to fashion and arts too. Beside \textit{carpet}, which we discussed in the introduction, such words include \textit{model}, \textit{string}, \textit{metal}, and so on. In the Bay Area, \textit{model} usually refers to mathematical models, and its closest words are \textit{freemium} (a pricing strategy), \textit{differentiator}, and \textit{hybrid}. The word in NYC is connected with fashion models, and closest to \textit{designer}, \textit{supermodel}, and \textit{stylist}. The popularity of music in NYC is clearly reflected in some regionally divergent words. For example, \textit{string} is associated with string instruments in NYC (closest words: \textit{strumming}, \textit{guitars}, and \textit{revolver}). Differently, in the Bay Area, the word is referred to as the data structure, which is closest to \textit{dict}, \textit{array}, and \textit{ascii}.  Similarly, \textit{metal} is highly related to heavy metal music in NYC, with closest words such as \textit{thrash} (thrash metal, a type of heavy metal music), \textit{sevenfold}, and \textit{metallica} (both are names of heavy metal bands). In the Bay Area, this word is interpreted as its literal meaning, and closest to \textit{brass}, \textit{rubber}, and \textit{aluminum}.

\section{ Conclusion \& Future Work}
In this paper, we have discussed divergent word interpretations among people. To detect these words, we propose an approach based on word embedding models, which quantifies differences in word interpretations. We apply the approach to two types of widely studied differences between people -- gender and regional differences. The discovered divergent words reveal the unique features of specific demographic groups of people, such as gender specific interests, cultural differences between regions, and so on. In the future, we would like to apply the approach to other types of human differences, such as age and occupation, and study the divergent word interpretations of these groups of people. Another possible direction is ``difference between differences''. It is interesting to study if some types of differences are more likely to lead to divergent word interpretations than others.

\balance{}

\bibliographystyle{aaai}

\bibliography{reference}

\end{document}